\pdfoutput=1

\documentclass[11pt]{article}

\usepackage[final]{acl}

\usepackage{times}
\usepackage{latexsym}
\usepackage{xcolor}
\usepackage{xspace}
\usepackage{comment}
\usepackage[normalem]{ulem}
\usepackage[utf8]{inputenc} 
\usepackage[T1]{fontenc}    
\usepackage{url}
\usepackage{tikz}
\usepackage{enumitem}
\usetikzlibrary{positioning}
\usepackage{array}
\usepackage{booktabs}
\usepackage{wrapfig}
\usepackage{caption} 
\usepackage{floatrow}
\usepackage{listings}
\usepackage{adjustbox}
\usepackage{footnote}
\usepackage{multirow}
\usepackage{amsmath}
\usepackage{breakcites}
\usepackage{blindtext}
\usepackage{svg}
\usepackage[most]{tcolorbox}
\usepackage{fnbreak}

\usepackage[T1]{fontenc}

\usepackage[utf8]{inputenc}

\usepackage{microtype}

\usepackage{inconsolata}

\usepackage{graphicx}

\usepackage{tabularx} 
\usepackage{booktabs} 
\usepackage{comment} 
\usepackage{wrapfig} 
\usepackage{calligra}

\usepackage{aurical}

\newcommand*{\escapeI}[1]{\texttt{\expandafter\string\csname #1\endcsname}}

\newenvironment{tight_itemize}{
\begin{itemize}
  \setlength{\itemsep}{0pt}
  \setlength{\parskip}{0pt}
}{\end{itemize}}

\makeatletter
\def\thickhline{%
  \noalign{\ifnum0=`}\fi\hrule \@height \thickarrayrulewidth \futurelet
   \reserved@a\@xthickhline}
\def\@xthickhline{\ifx\reserved@a\thickhline
               \vskip\doublerulesep
               \vskip-\thickarrayrulewidth
             \fi
      \ifnum0=`{\fi}}
\makeatother

\makeatletter
\newcommand\footnoteref[1]{\protected@xdef\@thefnmark{\ref{#1}}\@footnotemark}
\makeatother

\newlength{\thickarrayrulewidth}
\newcommand{\todo}[1]{{\color{red}\bf [TODO: #1]}\xspace}
\newcommand{\citereq}[1]{{\color{blue}\bf [CITE]}\xspace}

\newcommand{\ccderiv}{$\mathtt{CC_{dv}}$\xspace}
\newcommand{\ccmq}{$\mathtt{CC}$-$\mathtt{Medium}$\xspace}
\newcommand{\ccmhq}{$\mathtt{CC}$-$\mathtt{Medium}$-$\mathtt{High}$\xspace}
\newcommand{\cchq}{$\mathtt{CC}$-$\mathtt{High}$\xspace}
\newcommand{\phaseone}{$\mathcal{P}_1$\xspace}
\newcommand{\phasetwo}{$\mathcal{P}_2$\xspace}

\newcommand{\gainRO}{$3.4$\%\xspace}
\newcommand{\gainND}{$17$\%\xspace}
\newcommand{\gainROND}{$13.2$\%\xspace}

\newcommand{\poneboneptwobone}{\phaseone-$\mathtt{Blend1}$-\phasetwo-$\mathtt{Blend1}$\xspace}
\newcommand{\poneboneptwobtwo}{\phaseone-$\mathtt{Blend1}$-\phasetwo-$\mathtt{Blend2}$\xspace}
\newcommand{\poneboneptwobthree}{\phaseone-$\mathtt{Blend1}$-\phasetwo-$\mathtt{Blend3}$\xspace}
\newcommand{\poneboneptwobfour}{\phaseone-$\mathtt{Blend1}$-\phasetwo-$\mathtt{Blend4}$\xspace}
\newcommand{\poneboneptwofive}{\phaseone-$\mathtt{Blend1}$-\phasetwo-$\mathtt{Blend5}$\xspace}

\newcommand{\ponebfourptwobone}{\phaseone-$\mathtt{Blend4}$-\phasetwo-$\mathtt{Blend1}$\xspace}
\newcommand{\ponebfourptwobtwo}{\phaseone-$\mathtt{Blend4}$-\phasetwo-$\mathtt{Blend2}$\xspace}
\newcommand{\ponebfourptwobthree}{\phaseone-$\mathtt{Blend4}$-\phasetwo-$\mathtt{Blend3}$\xspace}
\newcommand{\ponebfourptwobfour}{\phaseone-$\mathtt{Blend4}$-\phasetwo-$\mathtt{Blend4}$\xspace}
\newcommand{\ponebfourptwobfive}{\phaseone-$\mathtt{Blend4}$-\phasetwo-$\mathtt{Blend5}$\xspace}

\newcommand{\ptwobone}{\phasetwo-$\mathtt{Blend1}$\xspace}
\newcommand{\ptwobsix}{\phasetwo-$\mathtt{Blend6}$\xspace}
\newcommand{\ponebfourptwobsix}{\phaseone-$\mathtt{Blend4}$-\phasetwo-$\mathtt{Blend6}$\xspace}

\title{Maximize Your Data's Potential: Enhancing LLM \\Accuracy with Two-Phase Pretraining 
} 

\author{Steven Y. Feng\thanks{equal contribution}$^{2}$\thanks{Work done during internship at NVIDIA}, Shrimai Prabhumoye$^{*}$$^{1,3}$, Kezhi Kong$^{1}$, Dan Su$^{1}$, \\ \textbf{Mostofa Patwary$^{1}$, Mohammad Shoeybi$^{1}$, Bryan Catanzaro$^{1}$} \\
NVIDIA$^{1}$, Stanford University$^{2}$, Boston University$^{3}$ \\
  \texttt{sprabhumoye@nvidia.com} \\}

\begin{document}
\maketitle
\begin{abstract}
    Pretraining large language models effectively requires strategic data selection, blending and ordering. 
    However, key details about data mixtures especially their scalability to longer token horizons and larger model sizes remain underexplored due to limited disclosure by model developers.
    To address this, we formalize the concept of \textit{two-phase pretraining} and conduct an extensive systematic study on how to select and mix data to maximize model accuracies for the two phases.
    Our findings illustrate that a two-phase approach for pretraining outperforms random data ordering and natural distribution of tokens by \gainRO and \gainND on average accuracies.
    We provide in-depth guidance on crafting optimal blends based on quality of the data source and the number of epochs to be seen.
    We propose to design blends using downsampled data at a smaller scale of 1T tokens and then demonstrate effective scaling of our approach to larger token horizon of 15T tokens and larger model size of 25B model size.
    These insights provide a series of steps practitioners can follow to design and scale their data blends.
   
\end{abstract}

\section{Introduction}

Large language models (LLM) are typically pretrained on large amounts of data in the order of billions (B) or trillions (T) of tokens derived from multiple data sources such as web crawl, books, papers, patents, mathematical and legal documents, and so forth~\cite{brown2020languagemodelsfewshotlearners,parmar2024nemotron415btechnicalreport,gemmateam2024gemmaopenmodelsbased,dubey2024llama,nvidia2024nemotron4340btechnicalreport}.
To develop a state-of-the-art model, it is critical to understand the nature of these data sources and to make informed decisions about optimal data blending (how different data sources are weighed during pretraining) 
and training strategies. 
These decisions typically involve running multiple large-scale experiments to empirically investigate the optimal training data blend(s) and ordering of data.


\begin{figure}[t]
    \centering
    \includegraphics[width=1.0\textwidth]{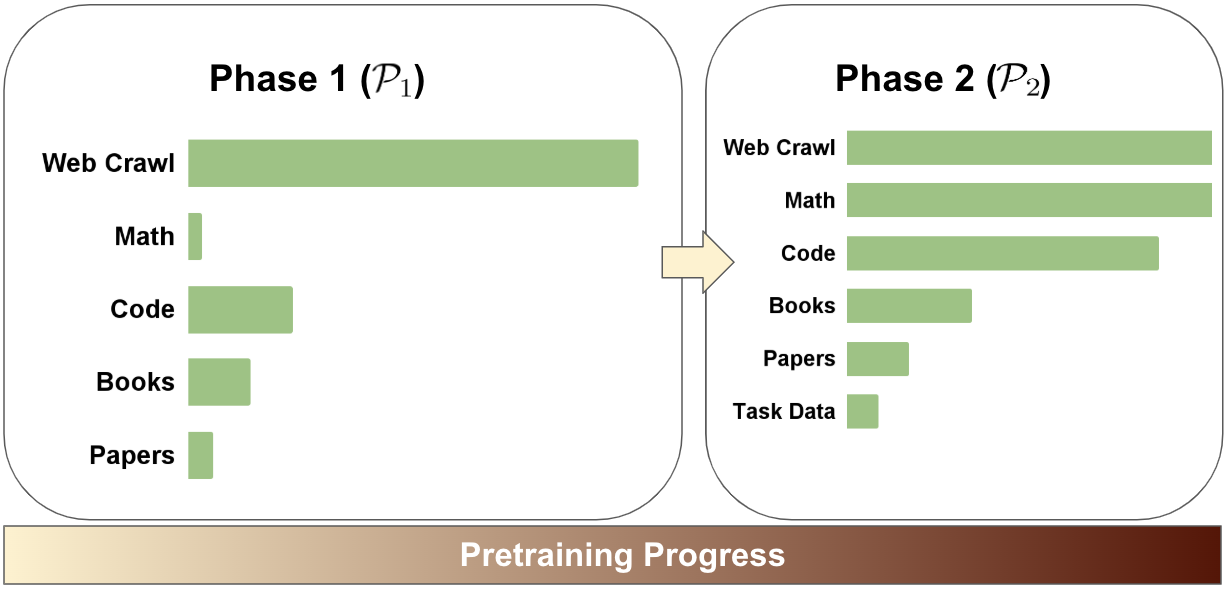}
    \caption{Diagram of our two phase training pipeline. Phase-1 blend encourages data diversity and phase-2 blend is focused on high quality datasets.
    } \label{fig:train_pipeline}
\end{figure}

Most advanced models~\cite{gpt4-openai2024gpt4technicalreport,llama3-dubey2024llama3herdmodels} do not divulge information on the data blends that are used, nor the ablation studies informing the data mixing and ordering decisions. 
Recent works \cite{domain-upsampling-blakeney2024doesdatasparkjoy,olmo-groeneveld-etal-2024-olmo,llama3-dubey2024llama3herdmodels,snowflake-arctic} provide high-level data blend information about a small portion of pretraining by encouraging the upsampling of certain domains towards the end.
In general, there exists a knowledge gap regarding how to craft and choose an optimal data blend(s) for the entire training process, and the generalizability of data blends and ordering strategies to larger token horizons and model sizes.

In this work, we address the above knowledge gap by understanding optimal data blends and ordering strategies for training LLM. We formalize and extensively explore a two-phase training approach (Figure~\ref{fig:train_pipeline}) that balances diversity and quality: phase-1 emphasizes diverse, high-quality web crawl data, while phase-2 focuses on high-quality data sources such as math, code, and wiki data.
Specifically, in this work we propose to use downsampled data to prototype and explore multiple blends at a smaller scale of 1T tokens.
We craft our blends based on quality of the data source and the number of epochs to be seen during pretraining.
We then demonstrate the effectiveness of our approach at a 15T token scale using the full data.

We evaluated on a comprehensive set of downstream tasking covering knowledge, reasoning, coding and math benchmarks.
Our experiments illustrate that a quality and epoch based blend is better than a blend based on natural distribution by \gainROND and the two-phase approach is better than random ordering of data (blend is based on quality and epochs) by an average of \gainRO across downstream tasks.
Furthermore, our results on downsampled data generalize across longer 15T token horizons on full data and larger model sizes, demonstrating the scalability and robustness of the two-phase approach. 
We also provide a fine-grained quality analysis of web crawl data, revealing optimal blending strategies to balance diversity and quality. 

We share and highlight a series of findings made to create blends and order in our two-phase approach.
Our main contributions are:



\vspace{-0.2cm}
\begin{tight_itemize}
\item Formalization and large-scale evaluation of the two-phase training approach for LLMs, with actionable strategies that enable effective LLM pretraining.
\item Improving the understanding of data selection and blending with quality-based and epoch-based analyses of data, including web crawl.
\item Demonstration of the scalability of blends using downsampled data at 1T to using full data at 15T tokens and larger model size of 25B. 
\end{tight_itemize}

\section{A Two-Phase Approach to Pretraining}

In this work, we explore a two-phased approach to pretraining: phase-1 (\phaseone) then phase-2 (\phasetwo).
Figure~\ref{fig:train_pipeline} demonstrates our two-phased approach.
In each phase, we explore different data blends based on the quality and number of epochs to be seen of a data source.
In phase-1 (\phaseone), we explore a general data distribution 
which consists of a mix of web crawl data, medium-quality data, and low amounts of high-quality data.
In phase-2 (\phasetwo), we explore a blend 
which includes task data and emphasizes high-quality datasets such as math, code, and high-quality web crawl (\S\ref{subsec:crawl_data_mix}).
As seen in Figure~\ref{fig:train_pipeline}, 
our model sees the first general  data blend during \phaseone for the majority of training, then a different data blend focused on high quality data during the shorter \phasetwo of training.

The steps to create blends for \phaseone and \phasetwo are: 1) Downsample a data source by a factor of $f$, 2) Estimate the quality of a data source (\S\ref{subsec:crawl_data_mix}), 3) Estimate the epochs to be seen in the whole pretraining (\S\ref{subsec:epoch_results}) and finally 4) distribute the epochs appropriately in \phaseone and \phasetwo (\S\ref{sec:data-blends}).
The downsampling factor $f$ is based on the final total token budget which we assume to be $15$T similar to \citet{llama3-dubey2024llama3herdmodels}.
Hence, for us $f=1/15$ i.e for each data source, the number of tokens available for pretraining is $1/15^{th}$ of the total token in that dataset.
Downsampling helps to observe the impact of epochs of datasets at a smaller scale of 1T tokens and then can be used to scale the blend to a longer token horizon of 15T tokens using the full data. 

\paragraph{Baselines:} Since our blends are based on quality and epoch based analyses of the data as well as the ordering of the data in the two phases, we consider the following two baselines: 1) Natural Distribution Blend (BASE-ND): This blend is based on ratio of the number of tokens available in each data source. The weight for each dataset is equal to the total number of tokens in that dataset divided by the sum of tokens available in all the datasets. This weighting is neither based on quality nor the epochs to be seen for the dataset. 2) Random Order Pretraining (BASE-RO): This blend is based on quality and epochs of each dataset but does not use two phases to train the model. The weight for each dataset here is the same as our two-phase approach but the order in which the the dataset is seen during pretraining is random.

\section{Experimental Setup}

\begin{table}[t]
\begin{center}
\scalebox{0.85}{\begin{tabular}{@{}lr@{}}
\toprule
\textbf{Data Domain} & \textbf{Tokens (B)} \\
\toprule
Web Crawl & 6244.3 \\
Math & 161.5\\
Wiki & 16.7 \\
Code & 760.3 \\
Books & 776.3 \\
Papers & 212.6 \\
\ccderiv & 348.3 \\
Multilingual & 1457.2 \\
Task Data & 6.6 \\
\bottomrule
\end{tabular}}
\end{center}
\caption[]{Tokens (billions) in each data domain.\label{tab:data-tokens}}
\end{table}

\subsection{Data Sources}
\label{subsec:data_sources}

Our pretraining corpus spans a vast range of text data sources that cover several domains, types of data, and languages. 
We broadly divide our datasets into the following categories and their token counts in billions is shown in Table~\ref{tab:data-tokens}.

\begin{tight_itemize}
    \item \textbf{Web Crawl:} Data derived from Common Crawl (CC). We discuss the quality of this data and how to blend it in detail in \S\ref{subsec:crawl_data_mix}.
    \item \textbf{High-Quality:} This includes datasets from more specialized and professional domains such as mathematics \cite{paster2024openwebmath,stackexchange}, code \cite{li2023starcoder}, and Wikipedia (wiki) 
    data. 
    \item \textbf{Medium-Quality:} Data derived from books \& patents,  
    papers \cite{gao2020pile800gbdatasetdiverse}, and Common Crawl derivatives (\ccderiv) such as OpenWebText \cite{Gokaslan2019OpenWeb}, BigScience \cite{bigscience}, Reddit \cite{baumgartner2020pushshiftredditdataset}, and CC-News. This category was determined by comparing this data to medium-quality crawl (see \S\ref{subsec:crawl_data_mix}). 
    \item \textbf{Multilingual:} Multilingual data (9 languages) derived from Wikipedia and Common Crawl.
    \item \textbf{Task Data:} This includes data used for supervised finetuning (SFT) during the alignment phase~\cite{toshniwal2024openmathinstruct,nvidia2024nemotron4340btechnicalreport}. We also include the FLAN collection~\cite{longpre2023flan}.
\end{tight_itemize}

\begin{table}[t]
\begin{center}
\scalebox{0.65}{\begin{tabular}{@{}llrrrrr@{}}
\toprule
\textbf{Category} & \textbf{Domain} & \textbf{Blend1} & \textbf{Blend2} & \textbf{Blend3}	& \textbf{Blend4} & \textbf{Blend5} \\
\toprule
Web Crawl & - & 65.0 & 65.0 & 58.0 & 59.0 & 70.0 \\
\midrule
\multirow{3}{*}{\parbox[c]{1.cm}{\centering High \\ Quality}} & Math & 1.9 & 1.9 & 1.9 & 2.9 & 1.9 \\
& Wiki & 0.1 & 0.1 & 0.1 & 0.1 & 0.1 \\
& Code & 15.0 & 8.0 & 15.0 & 20.0 & 13.0 \\
\midrule
\multirow{3}{*}{\parbox[c]{1.cm}{\centering Medium \\ Quality}} & Books & 5.5 & 9.0 & 9.0 & 5.5 & 4.5 \\
& Papers & 3.5 & 5.0 & 5.0 & 3.5 & 1.9 \\
& \ccderiv & 4.0 & 6.0 & 6.0 & 4.0 & 3.6 \\
\midrule
Multilingual & - & 5.0 & 5.0 & 5.0 & 5.0 & 5.0 \\
\bottomrule
\end{tabular}}
\end{center}
\caption[]{Phase-1 Blends (in \%)
\label{tab:phase1-blends}}
\vspace{-0.5em}
\end{table}

\subsection{Data Blends for Each Phase}
\label{sec:data-blends}

The final blends in \phaseone and \phasetwo are based on quality and epoch based ablations shown in \S\ref{subsec:crawl_data_mix} and \S\ref{subsec:epoch_results}. The insights from these studies are incorporated in Table~\ref{tab:phase1-blends} and \ref{tab:phase2-blends}.

In \phaseone, we encourage diversity in data by including a high percentage of web crawl data which consists of high, medium, and low-quality crawl.
We want to introduce a limited amount of high-quality data such as math, code, and wiki in \phaseone. 
In \phasetwo, the emphasis is primarily on high-quality datasets and only includes a limited amount of medium-quality data.
For example, in \phasetwo, we only use high-quality crawl instead of medium or low-quality (see \S\ref{subsec:crawl_data_mix}).

Table~\ref{tab:phase1-blends} details the five blends explored in \phaseone. These blends are designed to compare the proportion of high-level categories with each other. The difference between $\mathtt{Blend1}$ and $\mathtt{Blend2}$ is that $\mathtt{Blend2}$ has less code and more medium-quality datasets compared to $\mathtt{Blend1}$. $\mathtt{Blend3}$ has less web crawl and more medium-quality datasets compared to $\mathtt{Blend1}$. $\mathtt{Blend4}$ has less web crawl and more high-quality datasets compared to $\mathtt{Blend1}$. $\mathtt{Blend5}$ is designed to have majority web crawl at the cost of code and medium-quality data.

Table~\ref{tab:phase2-blends} outlines the five blends explored in \phasetwo.
In \phasetwo, we use more epochs and higher proportions of high-quality data such as high-quality web crawl, math, wiki, and code data.
$\mathtt{Blend3}$ has more code and less medium-quality datasets compared to $\mathtt{Blend1}$, and $\mathtt{Blend4}$ has more high-quality web crawl and less medium-quality datasets compared to $\mathtt{Blend1}$. 
$\mathtt{Blend2}$ has a more balanced distribution among the data categories, while $\mathtt{Blend5}$ upsamples math data more heavily. 

\begin{table}[t]
\begin{center}
\scalebox{0.65}{\begin{tabular}{@{}llrrrrr@{}}
\toprule
\textbf{Category} & \textbf{Domain} & \textbf{Blend1} & \textbf{Blend2} & \textbf{Blend3}	& \textbf{Blend4} & \textbf{Blend5} \\
\toprule
Web Crawl & - & 31.0 & 35.0 & 31.0 & 40.0 & 35.0 \\
\midrule
\multirow{3}{*}{\parbox[c]{1.cm}{\centering High \\ Quality}} & Math & 24.0 & 24.0 & 24.0 & 24.0 & 29.0 \\
& Wiki & 1.0 & 1.0 & 1.0 & 1.0 & 1.0 \\
& Code & 20.0 & 25.0 & 29.0 & 20.0 & 20.0 \\
\midrule
\multirow{3}{*}{\parbox[c]{1.cm}{\centering Medium \\ Quality}}
& Books & 8.0 & 4.0 & 4.0 & 4.0 & 4.0 \\
& Papers & 4.0 & 2.0 & 2.0 & 2.0 & 2.0 \\
& \ccderiv & 7.0 & 4.0 & 4.0 & 4.0 & 4.0 \\
\midrule
Multilingual & - & 3.7 & 3.7 & 3.7 & 3.7 & 3.7 \\
\midrule
Task Data & - & 1.3 & 1.3 & 1.3 & 1.3 & 1.3 \\
\bottomrule
\end{tabular}}
\end{center}
\caption[]{Phase-2 Blends (in \%) 
\label{tab:phase2-blends}}
\vspace{-0.5em}
\end{table}

\subsection{Model Specifications}
We experiment using the Megatron \cite{shoeybi2020megatronlmtrainingmultibillionparameter} model, an autoregressive causal left-to-right LLM, with the Tiktokenizer~\cite{tiktoken}.
We downsample all our data by factor $f=1/15$. 
Hence, only $1/15$ of the tokens shown in Table~\ref{tab:data-tokens} will be available for pretraining.
We perform all our investigations using an 8 billion parameter model trained on 1 trillion total tokens.
Furthermore, we test our two-phase approach by scaling along two dimensions: (1) we scale the token horizon to 1.7T tokens on a 8B model, and (2) we scale the parameters of the model to 25B and train on 1T tokens. 
Additionally, we train a 8B model on 15T tokens on full data (not downsampled) to observe if decisions made with downsampled data scales. Specifics on model architecture and hyperparameters are shared in Appendix \ref{app:detailed_model_specs}.

\subsection{Evaluation Suite}
\label{sec:eval-suite}

To comprehensively assess our models, we use various benchmarks that evaluate different capabilities. These can be broadly divided into the following 4 categories, of which we report the final averages. We assess 5-shot accuracy for MMLU \cite{hendrycks2021MMLU}, 0-shot accuracy\footnote{We use normalized accuracy for ARC-Easy, ARC-Challenge, PIQA, HellaSwag, and OpenBookQA.} for reasoning tasks:  CommonsenseQA \cite{talmor-etal-2019-commonsenseqa}, ARC-Easy \& Challenge \cite{ARC-clark2018thinksolvedquestionanswering}, PIQA \cite{PIQA-bisk2019piqareasoningphysicalcommonsense}, WinoGrande \cite{WinoGrande-sakaguchi2019winograndeadversarialwinogradschema}, HellaSwag \cite{zellers-etal-2019-hellaswag}, OpenBookQA \cite{openbookQA-mihaylov-etal-2018-suit}, RACE \cite{RACE-lai-etal-2017-race}, 0-shot accuracy for code benchmarks: HumanEval (+) \cite{HumanEval-chen2021evaluatinglargelanguagemodels} and MBPP (+) \cite{MBPP-austin2021programsynthesislargelanguage}, and 8-shot chain-of-thought (CoT) accuracy for GSM8K \cite{GSM8K-cobbe2021trainingverifierssolvemath}. We also report a final overall \textit{Avg.} for most results, which is an average over all individual evaluation tasks.

\begin{table}[t]
\begin{center}
\scalebox{0.8}{\begin{tabular}{@{}lrrrrr@{}}
\toprule
\textbf{Exp.} & \textbf{MMLU} & \textbf{Reason.} & \textbf{GSM8K} & \textbf{Code}	& \textbf{Avg.} \\
\toprule
BASE-ND & 49.78 & 56.48 & 19.64 & 24.96 & 45.17 \\
BASE-RO & 56.49 & 59.69 & 30.86 & 35.55 & 51.12 \\
Two-Phase & 56.28 & 60.34 & 40.33 & 38.33 & \textbf{52.86} \\
\bottomrule
\end{tabular}}
\end{center}
\caption[]{Comparison of our two-phase training approach with BASE-ND and BASE-RO.
\label{tab:rand-vs-two-phase-result}}
\end{table}

\section{Results for Two-Phase Pretraining}
\label{subsec:two-phase-results}

\begin{tcolorbox}[colframe=black!80, colback=gray!10, coltitle=white, title=Findings, fonttitle=\bfseries]
\begin{itemize}
    \item A two-phase approach for pretraining is effective.
    \item Phase-1 should focus on data diversity and phase-2 on high-quality data.
\end{itemize}
\end{tcolorbox}

We compare our best blends \ponebfourptwobone \footnote{see \S\ref{sec:finalizing-blends} Tab.~\ref{tab:phase2-results} on how we select this blend and \S\ref{subsec:pct-upsampling} Tab.~\ref{tab:data-upsampling-results-2} for the duration of \phasetwo for best results.} using two-phase training with two baselines: 1) BASE-ND: the weights are determined by the tokens available in each dataset and are not based on quality, and 2) BASE-RO: the weights for all the datasets are the same in this and \ponebfourptwobone.
The only difference is the order in which the data is presented during training (random or two-phased).
Table \ref{tab:rand-vs-two-phase-result} illustrates that using a quality and epoch based blend is on average \gainROND better than natural distribution blend (compare BASE-RO vs BASE-ND) across downstream tasks.
It also presents that using our two-phase training approach noticeably improves average accuracy by \gainRO compared to BASE-RO and \gainND compared to BASE-ND.
This empirically demonstrates that the strategy of two-phase training is useful and tasks such as code and math are sensitive to the ordering of high-quality data in the second phase. 

We scale our best blend \ponebfourptwobone to 15T tokens and use the full dataset to train a 8B model.
All the previous experiments are performed on downsampled data and 1T scale. 
This means that the number of epochs is constant in both the runs.
Table~\ref{tab:15T-result} 
shows that blends crafted at smaller scale can generalize to longer token budgets if the quality and epochs of the datasets are maintained at scale.
This shows the generalizability of our two-phase approach to pretraining as well as quality- and epoch-based approach to designing blends.

\begin{table}[t]
\begin{center}
\scalebox{0.8}{\begin{tabular}{@{}lrrrrr@{}}
\toprule
\textbf{Tok.} & \textbf{MMLU} & \textbf{Reason.} & \textbf{GSM8K} & \textbf{Code}	& \textbf{Avg.} \\
\toprule
1T & 56.28 & 60.34 & 40.33 & 38.33 & 52.86 \\
15T & 70.30 & 64.11 & 64.82 & 46.38 & 59.84 \\
\bottomrule
\end{tabular}}
\end{center}
\caption[]{Results of our two-phase training approach with \ponebfourptwobone for downsampled data at 1T and then complete data at 15T.
\label{tab:15T-result}}
\end{table}

\subsection{Determining Blends}
\label{sec:finalizing-blends}


\begin{figure}[t]
    \centering
    \includegraphics[width=0.9\textwidth]{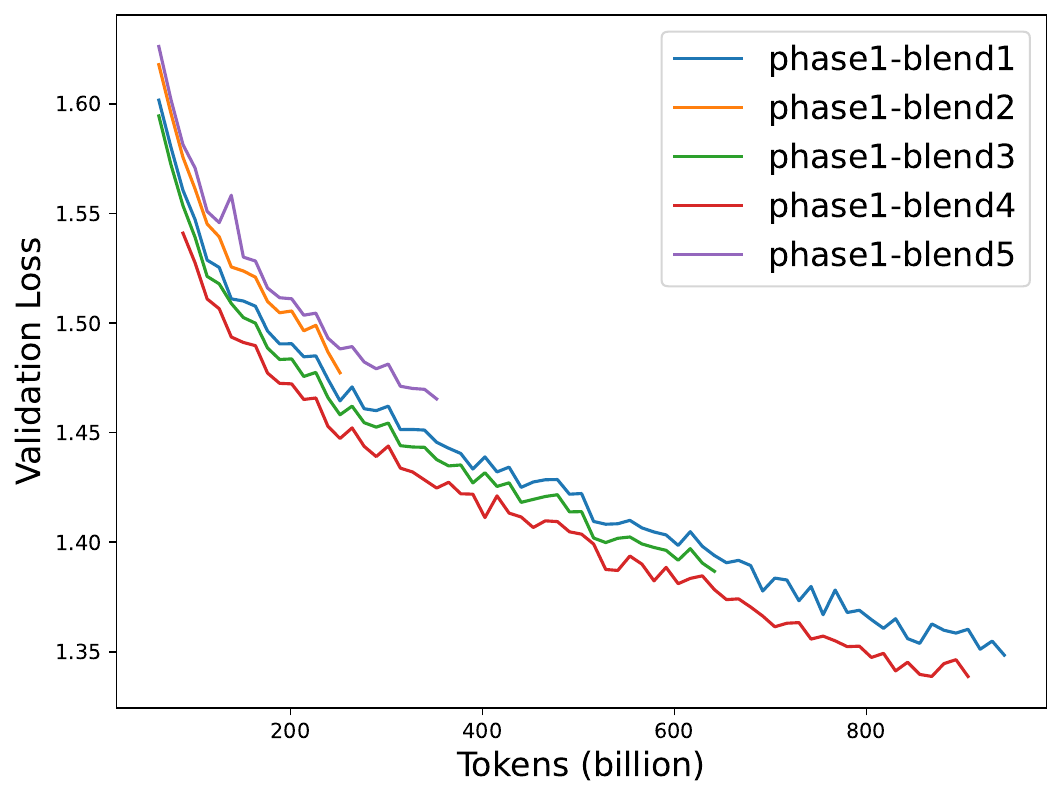}
    \caption{Phase-1 validation loss for different \phaseone blends.} \label{fig:phase1-val-loss}
\end{figure}

\begin{table}[t]
\begin{center}
\scalebox{0.7}{\begin{tabular}{@{}lcrrrrr@{}}
\toprule
\textbf{Exp.} & \textbf{Tokens} & \textbf{MMLU} & \textbf{Reason.} & \textbf{GSM8K} & \textbf{Code}	& \textbf{Avg.} \\
\toprule
\phaseone-$\mathtt{Blend1}$ &  200 & 34.72 & 52.83 & 6.14 & 16.43 & 38.81 \\
\phaseone-$\mathtt{Blend3}$ &  200 & 36.78 & 51.97 & 6.22 & 15.18 & 38.10 \\
\phaseone-$\mathtt{Blend4}$ &  200 & 38.70 & 53.81 & 11.30 & 18.21 & \textbf{40.48} \\
\midrule
\phaseone-$\mathtt{Blend1}$ &  250 & 42.51 & 54.52 & 7.51 & 16.14 & 40.35 \\
\phaseone-$\mathtt{Blend3}$ &  250 & 40.41 & 53.87 & 8.11 & 15.62 & 39.72 \\
\phaseone-$\mathtt{Blend4}$ &  250 & 42.76 & 54.99 & 10.16 & 19.29 & \textbf{41.66} \\
\midrule
\phaseone-$\mathtt{Blend1}$ &  629 & 51.93 & 57.83 & 14.94 & 22.97 & 45.28\\
\phaseone-$\mathtt{Blend3}$ &  629 & 52.44 & 57.74 & 15.39 & 22.43 & 45.15\\
\phaseone-$\mathtt{Blend4}$ &  629 & 52.78 & 58.11 & 18.27 & 24.24 & \textbf{46.07}\\
\bottomrule
\end{tabular}}
\end{center}
\caption[]{\phaseone results after various token counts (billions).
\label{tab:phase1-results}}
\end{table}


As discussed in \S\ref{sec:data-blends}, we explore five different blends for phase-1.\footnote{More detailed evaluation results of the major experiments in this section broken down by individual reasoning, MMLU, and code benchmarks and categories can be found in \S\ref{app:detailed_exp_results}.} 
We train an 8B model on downsampled data for 1T tokens for all five blends and eliminate blends based on a separately held-out validation split.
Fig.~\ref{fig:phase1-val-loss} illustrates the validation loss for all five blends.
As we can see, $\mathtt{Blend5}$ and $\mathtt{Blend2}$ had 2.8\% and 2.1\% higher validation loss, respectively, relative to $\mathtt{Blend4}$ at approx. 250B tokens.
Hence, we discontinue these two blends at that point.
Since, the validation loss of the remaining three blends was within a margin of 1\%, we periodically evaluate their accuracy on downstream tasks. 
Table~\ref{tab:phase1-results} shows the results of the remaining three phase-1 blends at various token counts.
At each token evaluation point -- 200B, 250B and 629B, we see that $\mathtt{Blend3}$ is consistently worse than the other two blends.
Hence, we eliminate this blend after 629B tokens of training. 
For this experiment, we switch from \phaseone to \phasetwo after $\approx$70\% of training, i.e the last 30\% of training is \phasetwo. 
In \S\ref{subsec:pct-upsampling}, we explore varying the percentage of \phasetwo.

Results in Table~\ref{tab:phase1-results} follow intuition since $\mathtt{Blend4}$ has the highest amount of high-quality data and is hence better than $\mathtt{Blend1}$ and $\mathtt{Blend3}$. $\mathtt{Blend3}$ has more medium-quality data at the cost of web crawl compared to $\mathtt{Blend1}$. 
This result confirms that books, papers, and \ccderiv are of medium-quality compared to our high-quality datasets and our web crawl blend. 

\begin{table}[t]
\begin{center}
\scalebox{0.68}{\begin{tabular}{@{}lrrrrr@{}}
\toprule
\textbf{Exp.} & \textbf{MMLU} & \textbf{Reason.} & \textbf{GSM8K} & \textbf{Code}	& \textbf{Avg.} \\
\toprule
\phaseone-$\mathtt{Blend1}$ & 55.00 & 59.12 & 20.09 & 23.56 & 46.76 \\
\phaseone-$\mathtt{Blend4}$ & 56.25 & 59.54 & 23.43 & 27.61 & 48.40 \\
\midrule
\poneboneptwobone & 56.04 & 60.04 & 37.00 & 36.19 & 51.88 \\
\poneboneptwobtwo & 55.88 & 60.15 & 36.85 & 35.89 & 51.84 \\
\poneboneptwobthree & 55.80 & 60.08 & 39.80 & 35.75 & 51.96 \\
\poneboneptwobfour & 56.15 & 60.26 & 36.85 & 36.30 & 51.88 \\
\poneboneptwofive & 56.49 & 60.41 & 36.92 & 34.40 & 51.65 \\
\ponebfourptwobone & 56.58 & 60.18 & 37.98 & 37.01 & \textbf{52.28} \\
\ponebfourptwobtwo & 56.89 & 60.00 & 36.62 & 36.97 & 52.10 \\
\ponebfourptwobthree & 56.10 & 60.08 & 39.27 & 35.15 & 51.78 \\
\ponebfourptwobfour & 57.03 & 59.98 & 36.85 & 36.30 & 51.93 \\
\ponebfourptwobfive & 56.92 & 60.35 & 38.29 & 34.56 & 51.77 \\
\bottomrule
\end{tabular}}
\end{center}
\caption[]{Evaluation results after \phasetwo of training. 
\label{tab:phase2-results}}
\end{table}

Finally, we explore five different blends of \phasetwo described in Table~\ref{tab:phase2-blends} in combination with \phaseone-$\mathtt{Blend1}$ and \phaseone-$\mathtt{Blend4}$.
Hence, we have ten different combinations of \phaseone and \phasetwo blends. 
Table~\ref{tab:phase2-results} shows the results on all ten combinations of blends.
We find that \ponebfourptwobone performs the best on average.
Table~\ref{tab:phase2-results} also presents the final results of \phaseone-$\mathtt{Blend1}$ and \phaseone-$\mathtt{Blend4}$ if only the \phaseone blend was continued for 1T tokens without ever switching to \phasetwo blends.
It shows that switching to any of the \phasetwo blends for training is better than continuing the \phaseone blends for all metrics.
We observe the largest absolute gains in GSM8K and code of 14.6\% and 9.4\%, respectively, for \ponebfourptwobone.

\begin{table}[t]
\begin{center}
\scalebox{0.6}{\begin{tabular}{@{}lcrrrrr@{}}
\toprule
\textbf{Exp.} & \textbf{Tok.} & \textbf{MMLU} & \textbf{Reason.} & \textbf{GSM8K} & \textbf{Code}	& \textbf{Avg.} \\
\toprule
\ponebfourptwobone & 1T & 56.58 & 60.18 & 37.98 & 37.01 & 52.28 \\
\ponebfourptwobone & 1.7T & 56.61 & 60.88 & 42.15 & 37.62 & 53.28 \\
\ponebfourptwobsix & 1.7T & 59.85 & 61.63 & 43.90 & 39.61 & \textbf{54.45} \\
\bottomrule
\end{tabular}}
\end{center}
\caption[]{
Scaling results for 1.7T tokens vs. 1T tokens, with and without high-quality data epoch adjustment.}
\label{tab:1.7T-results}
\end{table}

\subsection{Scaling}
\label{subsec:scaling_results}

\begin{tcolorbox}[colframe=black!80, colback=gray!10, coltitle=white, title=Findings, fonttitle=\bfseries]
\begin{itemize}
\item Two-phase approach is scalable and robust to token horizon and model scale.
\item Data blends need adjusting at longer token horizons based on epoch count to avoid high-quality data overexposure.
\end{itemize}
\end{tcolorbox}

We further explore scaling our best blend along two dimensions: (1) a longer token horizon of 1.7 trillion tokens and (2) larger model size of 25B parameters.
For a longer token horizon, we aim to assess whether the blend can be used as is or if adjustments are necessary to prevent overfitting (observed in \S\ref{subsec:epoch_results}).
Note that this is different from scaling to 15T token where we use the full data.
Here we still use the downsampled data and scale to 1.7T tokens and hence the epochs seen of each dataset would be higher.
Since high number of epochs of high-quality datasets are primarily seen in \phasetwo of pretraining, we create a new blend, \ptwobsix\footnote{We show comparison of $\mathtt{Blend1}$ and $\mathtt{Blend6}$ in Table~\ref{tab:phase2-blend6}.}, which is an epoch-adjusted version of \ptwobone to ensure that we do not see more than 8 epochs of certain high-quality data sources like math and task data. 
Table~\ref{tab:1.7T-results} shows the comparison of scaling 
from 1T to 1.7T total tokens.
We see that \ponebfourptwobsix is on average $2.2$\% better than \ponebfourptwobone, illustrating that we need to adjust our blends according to the epoch counts of high-quality data for optimal results.
Both the 1.7T models are better than 1T, demonstrating that we can still obtain higher downstream accuracies by training on more tokens, even if it means training on more than 8 epochs of high-quality data.

\begin{figure}[t]
    \centering
    \includegraphics[width=0.9\textwidth]{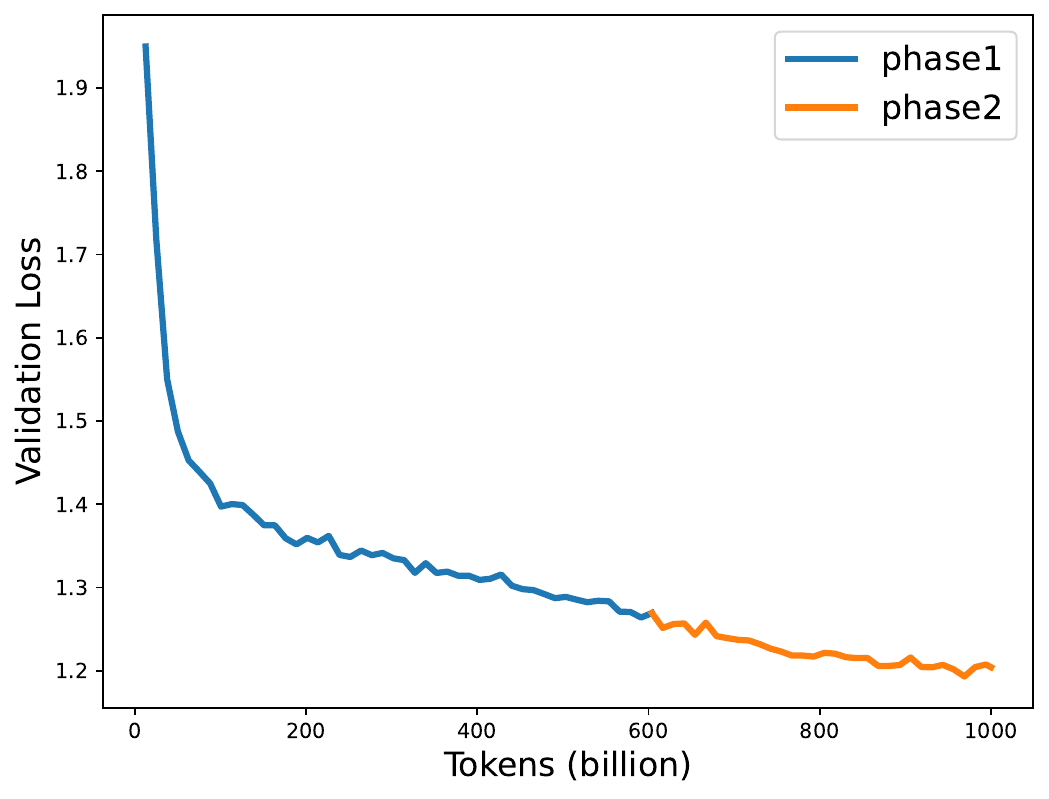}
    \caption{Validation loss for the 25B model using two-phase training with \ponebfourptwobone. \label{fig:25B-val-loss}}
\end{figure}

We also investigate if our best blend can scale to a larger model size.
Given the high number of epochs of high-quality data in \ponebfourptwobone, we also want to determine if a model with a larger capacity might memorize the data and overfit on it.
Figure~\ref{fig:25B-val-loss} shows that the validation loss is always decreasing for the 25B model, indicating that there is no overfitting with \ponebfourptwobone.
Table~\ref{tab:25B-results} shows results for the 25B model compared to the 8B model on this data blend combination. 
Understandably, the 25B model is substantially better across the board, demonstrating that the two-phase training approach and data blend combination can also scale to larger model sizes.

\begin{table}[t]
\begin{center}
\scalebox{0.8}{\begin{tabular}{@{}crrrrr@{}}
\toprule
\textbf{Model Size} & \textbf{MMLU} & \textbf{Reason.} & \textbf{GSM8K} & \textbf{Code}	& \textbf{Avg.} \\
\toprule
8B & 57.31 & 61.16 & 45.11 & 38.97 & 53.92 \\
25B & 65.97 & 63.29 & 59.14 & 45.57 & \textbf{58.47} \\
\bottomrule
\end{tabular}}
\end{center}
\caption[]{
Evaluation results for 8B vs. 25B parameter models, using the same blend: \ponebfourptwobone. Note that we use a maximum sequence length of 8192 (instead of 4096) for both models here.
\label{tab:25B-results}}
\end{table}

\section{Ablations}
\label{sec:ablations}

This section details quality-based data blending, epoch study and percentage of phase-2 to be conducted in the pretraining.
Additional fine-grained analyses of data blends and study on learning rate schedule to be used in phase-2 is shown in Appendix \S\ref{app:finegrained_upsampling_exps} and \S\ref{app:lr_schedule}.

\subsection{Quality-Based Data Blending}
\label{subsec:crawl_data_mix}

\begin{tcolorbox}[colframe=black!80, colback=gray!10, coltitle=white, title=Insights, fonttitle=\bfseries]
\begin{itemize}
\item Upsampling high quality and not using low quality CC data is most effective.
\item \ccderiv, papers and books are similar in quality to \ccmhq.
\end{itemize}
\end{tcolorbox}


The data blends of our two-phase approach are mainly based on the assessment of each data source's quality.
Hence, we carry out extensive experiments to find an optimal data blend for web crawl documents.
While previous work~\citep{dubey2024llama,yang2024qwen2,team2024gemma} mentions that web crawl documents like Common Crawl (CC) form a large majority of their pretraining data, none of them share a recipe on how to mix different slices of CC. 
Some recent work on constructing crawl-based pretraining datasets ~\citep{penedo2024fineweb,li2024datacomp} directly use the high quality crawl documents in pretraining but provide no specific data mixing strategy.
In this section, we provide comprehensive details on how to create a data blend for CC documents and use it effectively in our phase-1 and phase-2 of pretraining.
Additionally, we provide a quality assessment of other datasets like \ccderiv, papers, books and our high quality datasets. 
We compare them with medium, and high quality web crawl to position them optimally in our \phaseone and \phasetwo blends.

\begin{table}[t]
\centering
\resizebox{\textwidth}{!}{%
\begin{tabular}{ccrrrr}
\toprule
\textbf{Quality Label} & \textbf{Token} & $\mathtt{CC}$-$\mathtt{Blend1}$ & $\mathtt{CC}$-$\mathtt{Blend2}$ & $\mathtt{CC}$-$\mathtt{Blend3}$ & $\mathtt{CC}$-$\mathtt{Blend4}$  \\ 
\midrule
High & 35.96\% & 57.0 & 57.0 & 51.5 & 45.0 \\
Medium-High & 8.56\% & 25.0 & 25.0 & 23.5 & 20.0 \\
Medium & 34.25\% & 18.0 & 13.0 & \multirow{2}{*}{23.0} & \multirow{2}{*}{32.0} \\
Medium-Low & 15.41\% & 0.0 & 2.0 &  &  \\
Low & 5.82\% & 0.0 & 3.0 & 2.0 & 3.0 \\ 
\bottomrule
\end{tabular}%
}
\caption{CC blends (in \%) by quality. For $\mathtt{CC}$-$\mathtt{Blend3}$ and $\mathtt{CC}$-$\mathtt{Blend4}$, we merged the Medium and Medium-Low categories. \textit{Token} column refers to the the natural distribution of tokens, i.e. percentage of total CC data that belongs to each category.
}
\label{table: crawl_data_mixing_phase1}

\end{table}

\paragraph{Quality-Based Blending for Web Crawl:}

Each document in our web crawl data is classified into one of five quality categories: High, Medium-High, Medium, Medium-Low, and Low using the classifier from \citet{su2024nemotroncctransformingcommoncrawl}.

\begin{table}[]
\centering
\resizebox{0.88\textwidth}{!}{%
\begin{tabular}{lrrrrr}
\toprule
\textbf{Exp.} & \textbf{MMLU} & \textbf{Reason.} & \textbf{GSM8K} & \textbf{Code} & \textbf{Avg.} \\ 
\midrule
$\mathtt{CC}$-$\mathtt{Blend1}$ & 57.09 & 61.16 & 13.42 & 19.78 & 46.01 \\
$\mathtt{CC}$-$\mathtt{Blend2}$ & 56.69 & 61.77 & 14.18 & 19.56 & 45.11 \\
$\mathtt{CC}$-$\mathtt{Blend3}$ & 56.29 & 60.74 & 14.25 & 18.44 & 44.17 \\
$\mathtt{CC}$-$\mathtt{Blend4}$ & 55.73 & 60.57 & 14.31 & 18.50 & 44.06 \\ 
\bottomrule
\end{tabular}%
}
\caption{\phaseone results using our various CC blends.}
\label{table: crawl_data_phase1_results}
\end{table}

We investigate various blends using quality-based weighted sampling approach\footnote{We show comparison of quality-based blending with natural token distribution-based blend in \S\ref{app:quality-based-blend}.} for all of web crawl data from 99 CC snapshots for our phase-1 of pretraining.
The idea is to upsample high and medium-quality crawl documents while avoiding a high quantity of low-quality data.
The overall idea for the four web crawl blends in Table~\ref{table: crawl_data_mixing_phase1} is to iteratively decrease the percentage of tokens from High and Medium-High and increase the tokens in the lower categories.
The results in Table~\ref{table: crawl_data_phase1_results}\footnote{The average in this table is primarily based on reasoning tasks and MMLU because these blends do not have math or code data.} demonstrate that eliminating the tail-end of the web crawl data belonging to Medium-Low and Low quality categories is beneficial as opposed to to keeping them for diversity.
Based on these results, we choose $\mathtt{CC}$-$\mathtt{Blend1}$ as the final data blend for web crawl documents to be used in all our final \phaseone blends (\S\ref{subsec:two-phase-results} and Table \ref{tab:phase1-blends}).
For \phasetwo, we only use web crawl data that belongs to \textit{High}-quality category.

\begin{table}[t]
\begin{center}
\scalebox{0.75}{\begin{tabular}{@{}lrrrr@{}}
\toprule
\textbf{Dataset} & \textbf{MMLU} & \textbf{Reason.} & \textbf{GSM8K} & \textbf{Avg.} \\
\toprule
\ccmq & 52.86 & 56.00 & 18.04 & 42.30 \\
\ccmhq & 53.75 & 58.09 & 18.50 & 43.45 \\
\cchq & 55.82 & 59.65 & 20.85 & 45.44 \\
\midrule
High Quality & 54.53 & 58.51 & 24.11 & 45.72 \\
\ccderiv & 54.47 & 58.20 & 20.14 & 44.27 \\
Books & 55.36 & 58.93 & 18.50 & 44.26 \\
Papers & 54.55 & 58.65 & 19.41 & 44.20 \\
\bottomrule
\end{tabular}}
\end{center}
\caption[]{Results of different quality crawl and their comparison with other datasets. Since code data is not included in most of these experiments, we exclude code evaluation.
\label{tab:crawl++_papers_vs_crawl}}
\end{table}

\begin{figure}[t]
    \centering
    \includegraphics[width=0.95\textwidth]{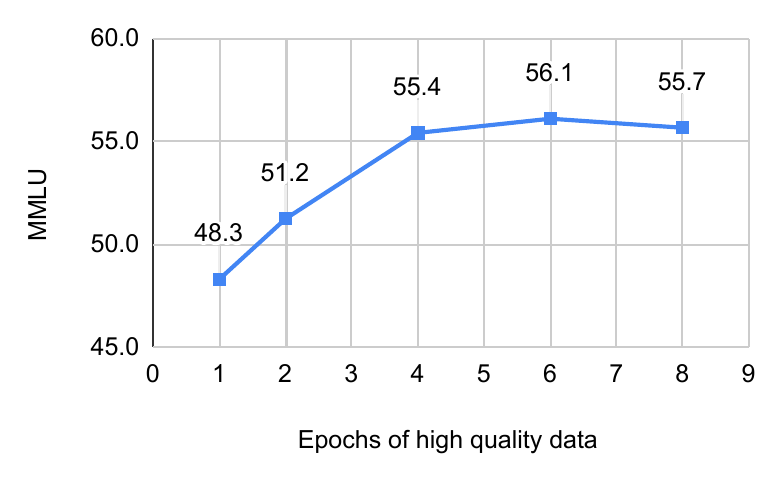}
    \\\vspace{-0.2\abovedisplayskip}
    \caption{MMLU accuracy (\%) vs. number of epochs of high-quality crawl in the data mix.}
    \label{fig:crawl_epoch}
\end{figure}

\paragraph{Quality Estimation of Other Datasets:}

We assess how our \ccderiv, papers, books and high quality datasets such as math, code and Wiki compare to \ccmq,\ccmhq and \cchq quality crawl data. 
We continue training the last checkpoint of \phaseone-$\mathtt{Blend4}$, for an additional 50B tokens using a data mix that consisted $66$\% of the data being tested, mixed with $34$\% of \cchq.

The results in Table \ref{tab:crawl++_papers_vs_crawl} show that \ccderiv, papers and books datasets have similar accuracies to \ccmhq on the majority of benchmarks, and lag behind \cchq. 
As such, we group them under the "medium-quality" data category for our experiments (see \S\ref{subsec:data_sources}). 
The high quality datasets have an average accuracy better than \cchq.

\subsection{Epoch-Based Analysis}
\label{subsec:epoch_results}

\begin{tcolorbox}[colframe=black!80, colback=gray!10, coltitle=white, title=Insights, fonttitle=\bfseries]
\begin{itemize}
\item We recommend 6 epochs of high-quality crawl and 8 epochs of math and task data for data mixing.
\end{itemize}
\end{tcolorbox}

We take the number of epochs of high quality datasets into account while creating our \phaseone and \phasetwo blends.
We experiment with different numbers of epochs for high-quality crawl, math, and task data.

Since, majority of web crawl is used in \phaseone, we pretrained an 8B model with 1T tokens, using different epochs of high-quality crawl tokens in the data mix, and evaluate each model's MMLU score. 
Note that we keep the overall percentage of web crawl the same in all the experiments.
As we can see in Figure~\ref{fig:crawl_epoch}, increasing the number of high-quality tokens increases the MMLU score until 6 epochs. 
We primarily present MMLU score because these experiments do not include high amount of math or code data.

\begin{table}[t]
\begin{center}
\scalebox{0.7}{\begin{tabular}{@{}lcrrrrr@{}}
\toprule
\textbf{Domain} & \textbf{Epochs} & \textbf{MMLU} & \textbf{Reason.} & \textbf{GSM8K} & \textbf{Code}	& \textbf{Avg.} \\
\toprule
Math & 1 & 57.06 & 60.51 & 36.47 & 33.55 & 51.49 \\
Math & 4 & 57.01 & 60.32 & 38.21 & 35.20 & 51.92 \\
Math & 8 & 56.58 & 60.18 & 37.98 & 37.01 & \textbf{52.28} \\
Math & 12 & 56.09 & 59.69 & 38.29 & 35.70 & 51.63 \\
\midrule
Task Data & 1 & 56.37 & 59.39 & 34.50 & 30.57 & 49.84 \\
Task Data & 4 & 56.57 & 59.46 & 40.18 & 35.27 & 51.53 \\
Task Data & 8 & 56.58 & 60.18 & 37.98 & 37.01 & \textbf{52.28} \\
Task Data & 12 & 56.77 & 59.96 & 38.44 & 36.04 & 51.93 \\
\bottomrule
\end{tabular}}
\end{center}
\caption[]{Results of varying the number of epochs of math and task data during \phasetwo of training.
\label{tab:epoch-results}}
\end{table}

Since, majority of math and task-data is seen in \phasetwo, Table~\ref{tab:epoch-results} presents results for different numbers of epochs for them in \phasetwo. 
It shows that $\approx$ 8 epochs of math is a good balance while not sacrificing accuracy on MMLU and reasoning. 
For task data, all metrics generally improve with more epochs, although there appears to be diminishing returns on several past epoch 8. 
Note that 8 epochs of both math and task data corresponds to our best \ponebfourptwobone blend combo from \S\ref{subsec:two-phase-results}.





\subsection{Optimal Duration of Phase-2}
\label{subsec:pct-upsampling}

\begin{tcolorbox}[colframe=black!80, colback=gray!10, coltitle=white, title=Insights, fonttitle=\bfseries]
\begin{itemize}
\item Pretraining with the \phasetwo blend for the final 40\% gives the best results.
\end{itemize}
\end{tcolorbox}

We investigate the percentage of phase-2 to use in the whole pretraining regime. 
We experiment with 0 to 50\% of \phasetwo in the whole of pretraining.
The longer the duration of \phasetwo, the shorter the duration of \phaseone.
We use the \ponebfourptwobone blend combination that we found best in \S\ref{subsec:two-phase-results}.

Table \ref{tab:data-upsampling-results-2} illustrates that a higher percentage of \phasetwo until 40\% is better overall, especially in math and code.
Going above this, e.g. to \phasetwo as 50\% of training, downstream accuracies start to degrade across the board, potentially due to overfitting. 


\begin{table}[t]
\begin{center}
\scalebox{0.75}{\begin{tabular}{@{}lrrrrr@{}}
\toprule
\textbf{\phasetwo \%} & \textbf{MMLU} & \textbf{Reason.} & \textbf{GSM8K} & \textbf{Code}	& \textbf{Avg.} \\
\toprule
$0$ & 56.10 & 61.81 & 16.60 & 16.22 & 37.68 \\
$10$ & 56.52 & 59.70 & 33.13 & 32.55 & 50.48 \\
$20$ & 56.54 & 59.93 & 40.16 & 34.29 & 51.58 \\
$30$ & 56.58 & 60.18 & 37.98 & 37.01 & 52.28 \\
$40$ & 56.28 & 60.34 & 40.33 & 38.33 & \textbf{52.86} \\
$50$ & 55.94 & 59.82 & 37.68 & 36.86 & 51.96 \\
\bottomrule
\end{tabular}}
\end{center}
\caption[]{Results of different durations of \phasetwo using \ponebfourptwobone. 
\label{tab:data-upsampling-results-2}}
\end{table}

\section{Related Work}


Selecting and structuring pretraining datasets is important to improve model generalization and efficiency.
\citet{llama3-dubey2024llama3herdmodels} emphasize openness and accessibility of models, \citet{together2023redpajama,soldaini-etal-2024-dolma} assemble an open corpus of trillions of tokens for large-scale training, and \citet{olmo-groeneveld-etal-2024-olmo} release a truly Open Language Model, including its framework, training data, and code. 
Studies such as \citet{li2024datacomplm,fineweb-penedo2024the} demonstrating that refined data selection impacts model accuracies more significantly than simply the quantity of data. 
But these studies are primarily aimed at CC data and they do not suggest any data mixing strategies for pretraining.
\citet{parmar-etal-2024-data} provide a systematic approach to building effective LLM pretraining datasets with ablations on data attributes, and existing curation, selection, and sampling methods. 
In our work, we provide a systematic approach to craft data blends and to order the data in pretraining.

Strategic weighting and timing of data usage can also noticeably impact model accuracies. 
Techniques like domain upsampling \cite{domain-upsampling-blakeney2024doesdatasparkjoy,llama3-dubey2024llama3herdmodels} towards the end of training have been shown to be effective.
\citet{snowflake-arctic,olmo-groeneveld-etal-2024-olmo} provide details about high level blends for their pretraining process.
In contrast, our work provides fine grained details about the data blend creation process along with actionable steps that model developers can use to develop data blends and order.
Prior work \cite{shen2023slimpajama,longpre2024pretrainer,pmlr-v162-mindermann22a,doremi-NEURIPS2023_dcba6be9,xie2023data,shao2024balanced} investigates optimizing data mixtures based on clustering methods, manually designed domain composition weights, proxy models or reference models to determine data composition weights and sample-level data selection. 
Our work primarily focuses on data ordering and scaling of data blends in pretraining and can be used in conjunction with other data sampling techniques.

Curriculum learning approaches inspired by human learning offer an ordered way to introduce data gradually to enhance model learning. 
\citet{martinez-etal-2023-climb,curriculum-survey,feng-etal-2024-child} investigate cognitively-motivated curriculum-based training including vocabulary, and objective curricula, and outline and the challenges and potential solutions for designing effective curricula. 
Our work shows that ordering of data based on quality in pretraining LLMs has a significant impact of downstream accuracies.

\section{Conclusion}
In conclusion, through extensive experiments, we demonstrate the effectiveness of a two-phase pretraining approach for LLM. 
For the initial training phase, a more general data distribution consisting of mainly of web crawl proves most effective, while phase two benefits from a comprehensive data blend, with additional focus on math, code, and task data. Phase-two for the last $\approx$40\% of training yields the best results, and over-extending it leads to diminishing returns. Increasing model size and token horizon further enhances accuracy, demonstrating the scalability of our approach. Importantly, we also show that considering both the quality of the data (including web crawl) and the number of epochs of each data source is crucial to attain optimal results and prevent overfitting.

\section*{Limitations}


Some limitations of our work include our present suite of models and evaluation benchmarks. 
We can extend our work and show the effectiveness of two-phase pretraining approach on more LLM architectures such as Mamba~\cite{gu2023mamba}, other hybrid SSM based architectures~\cite{glorioso2024zamba,lieber2024jamba} and mixture of experts~\cite{shazeer2017outrageously}. 
While our evaluation benchmarks are quite comprehensive, we could potentially expand to an even broader range of evaluations, including nuanced domain-specific or interactive tasks, or more theory of mind and developmental psychology-inspired benchmarks. 
This includes assessing capabilities such as analogical reasoning \cite{webb2023emergentanalogicalreasoninglarge}.
Further, our scaling experiments could be expanded. Scaling up to hundreds of billions of parameters or significantly longer training may yield additional insights. Lastly, while our work focuses on two-phase training and shows its efficacy, we can potentially investigate multi-phase training, and the impact of the order of the phases. However, we believe this is more suited for future work. Overall, these are directions to potentially improve and expand upon our work. 
Despite these potential limitations, we feel that our current work is an insightful and useful contribution to the research community.

\section*{Ethical Considerations}


Our research uses publicly available and commonly used datasets in LLM development. 
These sources, including Common Crawl, Wikipedia, and code repositories, are widely adopted in the research community. We examined the quality and origins of our data, prioritizing high-quality, domain-relevant data sources to improve LLM capabilities in a responsible manner. However, web crawl data may inherently contain biases or inappropriate content despite filtering efforts. We used established data cleaning and quality assurance procedures but acknowledge that potential biases may persist and impact model behavior in certain circumstances.


We recognize that scaling models and exploring data blending strategies require significant computational resources, which may raise environmental concerns. To mitigate this, we focused on efficient training strategies, such as two-phase training, to improve accuracy without excessively increasing resource usage. Future studies could benefit from exploring energy-efficient training methods to further minimize the environmental impact. 


Our models, data blends, and accompanying publication are intended solely for research purposes, with no intended real-world application without additional safety evaluations. We caution against deploying models based on our methods without thorough testing, as they may carry unknown risks, particularly when applied to tasks involving sensitive or personal information. Our work aims to advance the understanding of LLM training strategies, and we feel that it is an important contribution to the research community. We encourage researchers to expand upon our work while further investigating the ethical and societal implications of LLM.



\bibliography{custom}

\newpage
\appendix

\section{Model Specifications}
\label{app:detailed_model_specs}

We use RoPE position embeddings \cite{rope-paper}, RMSNorm layer normalization \cite{RMSNorm}, with Grouped Query Attention \cite{ainslie-etal-2023-gqa}. The maximum sequence length is 4096. We use a global batch size of 1536, 
and the Adam optimizer \cite{kingma2017adammethodstochasticoptimization} with $\beta=(0.9,0.95)$ and $\epsilon=1e$-$08$. 

\phaseone training uses cosine $LR$ decay with an initial $LR$ of $3e$-$4$ and targeted to reach a min-$LR$ of $3e$-$6$ at the end of the full training run (both phases). 
We start \phasetwo with the intermediate $LR$ reached at the end of \phaseone, and anneal using cosine $LR$ decay to $3e$-$6$ (\S\ref{app:lr_schedule}). 
Our experiments are run using up to 1024 NVIDIA H100 GPUs.

\section{Detailed Two-Phase Pretraining Results (Reasoning, MMLU, Code)}
\label{app:detailed_exp_results}

Tables \ref{tab:P2-eval-results-reasoning} to \ref{tab:model-scaling-eval-results-MMLU-code} contain detailed evaluation results of the major experiments reported in \S\ref{subsec:two-phase-results} for reasoning, MMLU, and code, broken down by individual categories and benchmarks. They correspond to the results found in Tables \ref{tab:phase2-results} to \ref{tab:25B-results} in \S\ref{subsec:two-phase-results}.

Table~\ref{tab:phase2-blend6} shows the comparison of $\mathtt{Blend1}$ and $\mathtt{Blend6}$ used in scaling experiments in Table~\ref{tab:1.7T-results}.
If we use the same $\mathtt{Blend1}$ as is and train for more number of tokens (1.7T) then the number of epochs seen of each dataset would be higher compared to 1T training.

\begin{table}[t]
\begin{center}
\scalebox{0.65}{\begin{tabular}{@{}llrr@{}}
\toprule
\textbf{Category} & \textbf{Domain} & \textbf{Blend1} & \textbf{Blend6} \\
\toprule
Web Crawl & - & 31.0 & 49.0 \\
\midrule
\multirow{3}{*}{High Quality} & Math & 24.0 &  14.4 \\
& Wiki & 1.0 & 0.6 \\
& Code & 20.0 & 12.0 \\
\midrule
\multirow{3}{*}{Medium Quality} & Books & 8.0 & 8.0 \\
& Papers & 4.0 & 4.0 \\
& \ccderiv & 7.0 & 7.0 \\
\midrule
Multilingual & - & 3.7 & 4.2 \\
\midrule
Task Data & - & 1.3 & 0.8\\
\bottomrule
\end{tabular}}
\end{center}
\caption[]{Comparison of $\mathtt{Blend1}$ and $\mathtt{Blend6}$ (in \%) used for scaling the token budget in \S\ref{subsec:scaling_results} and Table \ref{tab:1.7T-results}.
\label{tab:phase2-blend6}}
\end{table}

\section{Details of Quality-Based Data Blend}
\label{app:quality-based-blend}

\begin{table}[h]
\small
\centering
\resizebox{0.6\textwidth}{!}{%
\begin{tabular}{lrr}
\toprule
\textbf{Quality Label} & \textbf{ND} & \textbf{WS} \\
\midrule
High & 0.01 & 0.04 \\
Medium-High & 1.08 & 6.42 \\
Medium & 7.01 & 41.83 \\
Medium-Low & 26.46 & 25.09 \\
Low & 64.44 & 0.00 \\
\bottomrule
\end{tabular}%
}
\caption{Data blends for CC Quality estimation experiment. The overall percentage of the Common Crawl Snapshots in our experiments is fixed at 73.3\%.}
\label{table:crawl_data_mixing_percent}
\end{table}

We first compare a baseline blend (ND) which uses the natural distribution of tokens with a smartly constructed weighted sampling blend (WS).
ND is based on the number of tokens that belong in each category as opposed to utilizing the quality label 
i.e. if 59\% of the tokens belong to \textit{Low} then 59\% of tokens seen during pretraining would belong to \textit{Low}.
We then create a data blend (WS) based on weighted sampling of high and medium-quality tokens.
The idea is to upsample high and medium-quality crawl documents and not use the low-quality data at all.
Table~\ref{table:crawl_data_mixing_percent} shows the token percentages that belong to each of the five quality labels for both ND and WS blends.
Table~\ref{table: crawl_data_mixing_comp} illustrates the results of the two models trained on the ND and WS data blends of web crawl, respectively. 
We see that our data blend (WS) outperforms on most of the evaluation tasks by a large margin, and the improvement on MMLU is substantial. 

\begin{table}[]
\centering
\resizebox{\textwidth}{!}{%
\begin{tabular}{lcccc}
\toprule
\textbf{Data Mixing} & \textbf{MMLU} & \textbf{Reason.} & \textbf{GSM8K} & \textbf{Code} \\
\midrule
ND & 42.94 & 59.40 & 8.11 & 19.25 \\
WS & 56.10 & 61.60 & 11.98 & 17.41 \\
\bottomrule
\end{tabular}%
}
\caption{Our \textit{WS}: weighted sampling data mixing method outperforms the \textit{ND}: natural distribution method.}
\label{table: crawl_data_mixing_comp}
\end{table}

\section{Finegrained \phasetwo Blend Experiments}
\label{app:finegrained_upsampling_exps}


We investigate fine-grained \phasetwo blends to determine the optimal blend.
For these experiments, we use a model trained on a \phaseone blend for 900B tokens (10\% \phasetwo duration), with a linear $LR$ decay to $0$. 

\paragraph{Crawl, Math, \& Code:} We investigate different percentages of high-quality crawl, math, and code data as shown in Table~\ref{tab:phase2-finegrained-blends}. 
Table \ref{tab:phase2-finegrained-results} demonstrates that a higher amount of math data (i.e. 30\%) helps across the board. 
However, code data results are mixed, as too much code without enough math ($\mathtt{CMC}$-$\mathtt{B1}$) seems to hurt all non-code metrics. Comparing ($\mathtt{CMC}$-$\mathtt{B2}$) vs. ($\mathtt{CMC}$-$\mathtt{B3}$), more than 15\% code does not add as much value, as gains saturate. Trading off crawl data for more code data also slightly hurts MMLU. 
As such, we decide that a final blend consisting of a higher amount of crawl and math with a moderate amount of code seems best overall. This corresponds to $\mathtt{CMC}$-$\mathtt{B3}$ in Table \ref{tab:phase2-finegrained-blends}, which consists of 30\% crawl, 33\% math, and 15\% code. 

\paragraph{Task Data:} Second, we investigate the inclusion of task data. Specifically, adding FLAN and synthetically-generated GSM8K-train data (similar to data augmentation approaches \cite{feng-etal-2021-survey}) to the $\mathtt{CMC}$-$\mathtt{B3}$ blend. 
Our FLAN data consists of a mixture of normal FLAN and FLAN-CoT (chain-of-thought) data. 
We compare 10 and 20 epochs of FLAN. 
These blends can be found in Table \ref{tab:phase2-finegrained-blends-pt2}, with the results in Table \ref{tab:phase2-finegrained-results}. We can see that including synthetic GSM8K-train and FLAN data noticeably improves GSM8K scores while not detrimenting the other benchmarks. In fact, FLAN data also helps further improve MMLU and reasoning. 20 epochs of FLAN seems better than 10 epochs overall. Hence, including task data for \phasetwo of training seems to be a good idea.

\paragraph{All Data Mixture:} Lastly, we investigate a final \phasetwo data mixture which is a combination of all the data sources we tried, including FLAN, GSM8K, and relatively higher amounts of math and code data. For this experiment, we use 30\% upsampling with $LR$ cosine decay to $3e-6$. This blend can be found in Table \ref{tab:phase2-finegrained-blends-pt2}, with the results at the bottom of Table \ref{tab:phase2-finegrained-results}.\footnote{The $\mathtt{CMC}$-$\mathtt{Blend3}$-$30\%$ result at the bottom of Table \ref{tab:phase2-finegrained-results} is also using 30\% upsampling with $LR$ cosine decay to $3e-6$.} We find that mixing all data sources 
helps greatly with GSM8K, noticeably with coding and reasoning, while retaining accuracy on MMLU. Hence, the final \phasetwo blends we investigate in \S\ref{subsec:two-phase-results} (Table \ref{tab:phase2-blends}) are motivated by these ablations -- they are blends of all data sources, including task data, with higher proportions of math and code.

\section{Annealing Learning Rate Schedule}
\label{app:lr_schedule}

We investigate different learning rate ($LR$) schedules for phase \phasetwo. 
Specifically, using the same \phasetwo blend for a 10\% duration, 
we try different $LR$ strategies. 
We compare cosine vs. linear $LR$ decay functions, and also compare decaying to a final $LR$ of $0$ vs. $3e$-$6$ (1\% of the original \phaseone starting $LR$ of $3e$-$4$). 
Not decaying $LR$ entirely to 0 leaves room for post-training, which is likely preferable. 

As seen in Table \ref{tab:LR-anneal-results}, there is a negligible difference between linear and cosine $LR$ decay, so we choose cosine decay for consistency with \phaseone.
We also see that $LR$ decay to $3e$-$6$ is comparable to decaying all the way to $0$, while leaving room for post-training. 
Hence, our final chosen annealing strategy is cosine $LR$ decay to $3e$-$6$, which we use for our final two-phase experiments in \S\ref{subsec:two-phase-results}.

\begin{table*}[t]
\begin{center}
\scalebox{0.66}{\begin{tabular}{@{}lccccccccc@{}}
\toprule
\textbf{Exp.} & \textbf{ARC-Easy} & \textbf{ARC-Challenge} & \textbf{RACE} & \textbf{PIQA} & \textbf{WinoGrande} & \textbf{HellaSwag} & \textbf{OpenBookQA} & \textbf{CommonsenseQA} & \textbf{Avg.}\\
\toprule
\phaseone-$\mathtt{Blend1}$ & 75.97 & 51.19 & 36.36 & 80.96 & 67.64 & 76.23 & 44.00 & 53.07 & 59.12 \\
\phaseone-$\mathtt{Blend4}$ & 77.23 & 53.24 & 36.46 & 80.47 & 68.35 & 76.48 & 44.40 & 53.15 & 59.54 \\
\midrule
\poneboneptwobone & 78.32 & 51.54 & 36.75 & 79.76 & 66.54 & 76.44 & 43.80 & 61.67 & 60.04\\
\poneboneptwobtwo & 78.79 & 53.07 & 35.69 & 80.79 & 67.09 & 76.52 & 43.40 & 61.18 & 60.15\\
\poneboneptwobthree & 79.29 & 53.16 & 36.27 & 79.76 & 66.77 & 76.43 & 42.80 & 61.43 & 60.08 \\
\poneboneptwobfour & 78.37 & 52.99 & 36.65 & 80.30 & 66.93 & 76.67 & 43.80 & 61.92 & 60.26 \\
\poneboneptwofive & 79.21 & 52.56 & 36.75 & 80.63 & 67.25 & 76.64 & 44.00 & 61.83 & 60.41 \\
\textbf{\ponebfourptwobone} & 80.30 & 54.95 & 35.50 & 79.98 & 68.35 & 76.55 & 43.80 & 56.59 & 60.18 \\
\ponebfourptwobtwo & 79.92 & 54.10 & 35.50 & 80.20 & 67.96 & 76.75 & 43.80 & 56.35 & 60.00 \\
\ponebfourptwobthree & 79.92 & 54.27 & 35.12 & 80.20 & 67.56 & 76.39 & 44.20 & 57.08 & 60.08 \\
\ponebfourptwobfour & 79.92 & 54.27 & 35.89 & 80.47 & 67.25 & 76.79 & 43.80 & 56.18 & 59.98 \\
\ponebfourptwobfive & 79.29 & 54.44 & 37.03 & 79.98 & 67.80 & 76.92 & 44.80 & 57.41 & 60.35 \\
\bottomrule
\end{tabular}}
\end{center}
\caption[]{Final reasoning evaluation results after \phasetwo of training, broken down by individual benchmark. Corresponds to Table \ref{tab:phase2-results} in \S\ref{subsec:two-phase-results}.
\label{tab:P2-eval-results-reasoning}}
\end{table*}

\begin{table*}[t]
\begin{center}
\scalebox{0.70}{\begin{tabular}{@{}lcccccccccc@{}}
\toprule
\textbf{Exp.} & \multicolumn{5}{c}{\textbf{MMLU}} & \multicolumn{5}{c}{\textbf{Code}} \\
\cmidrule(lr){2-6} \cmidrule(lr){7-11}
 & \textbf{STEM} & \textbf{Humanities} & \textbf{Social Sciences} & \textbf{Others} & \textbf{Avg.} & \textbf{HumanEval} & \textbf{HumanEval+} & \textbf{MBPP} & \textbf{MBPP+} & \textbf{Avg.} \\
\toprule
\phaseone-$\mathtt{Blend1}$ & 45.61 & 50.24 & 65.29 & 61.54 & 55.00 & 18.90 & 13.41 & 31.52 & 30.42 & 23.56 \\
\phaseone-$\mathtt{Blend4}$ & 48.30 & 49.88 & 67.53 & 62.79 & 56.25 & 18.90 & 16.46 & 42.80 & 32.28 & 27.61 \\
\midrule
\poneboneptwobone & 47.57 & 51.12 & 65.88 & 62.34 & 56.04 & 32.32 & 27.44 & 42.41 & 42.59 & 36.19 \\
\poneboneptwobtwo & 48.65 & 50.41 & 65.78 & 61.67 & 55.88 & 31.71 & 26.83 & 42.41 & 42.59 & 35.89 \\
\poneboneptwobthree & 47.61 & 50.69 & 65.78 & 61.96 & 55.80 & 31.10 & 25.61 & 43.97 & 42.33 & 35.75 \\
\poneboneptwobfour & 48.11 & 50.84 & 65.81 & 62.76 & 56.15 & 28.66 & 25.61 & 42.80 & 42.86 & 35.59 \\
\poneboneptwofive & 48.94 & 51.41 & 66.14 & 62.28 & 56.49 & 28.66 & 23.78 & 42.02 & 43.12 & 34.40 \\
\ponebfourptwobone & 49.29 & 50.35 & 67.44 & 62.66 & 56.58 & 31.10 & 24.39 & 49.42 & 43.12 & 37.01 \\
\ponebfourptwobtwo & 49.44 & 50.92 & 67.47 & 62.99 & 56.89 & 30.49 & 25.00 & 49.81 & 42.59 & 36.97 \\
\ponebfourptwobthree & 49.19 & 49.37 & 66.88 & 62.60 & 56.10 & 27.44 & 20.73 & 48.25 & 44.18 & 36.15 \\
\ponebfourptwobfour & 49.16 & 50.84 & 68.35 & 63.18 & 57.03 & 31.71 & 23.17 & 47.47 & 42.86 & 36.30 \\
\ponebfourptwobfive & 49.38 & 50.86 & 68.18 & 62.60 & 56.92 & 28.66 & 20.12 & 45.53 & 43.92 & 34.56 \\
\bottomrule
\end{tabular}}
\end{center}
\caption[]{Final MMLU and code evaluation results after \phasetwo of training, broken down by individual category/benchmark. Corresponds to Table \ref{tab:phase2-results} in \S\ref{subsec:two-phase-results}.
\label{tab:P2-eval-results-MMLU-code}}
\end{table*}

\begin{table*}[t]
\begin{center}
\scalebox{0.71}{\begin{tabular}{@{}lccccccccc@{}}
\toprule
\textbf{Exp.} & \textbf{ARC-Easy} & \textbf{ARC-Challenge} & \textbf{RACE} & \textbf{PIQA} & \textbf{WinoGrande} & \textbf{HellaSwag} & \textbf{OpenBookQA} & \textbf{CommonsenseQA} & \textbf{Avg.}\\
\toprule
BASE & 78.75 & 53.84 & 35.69 & 80.30 & 68.51 & 76.30 & 45.40 & 51.68 & 59.69 \\
Two-Phase & 80.30 & 54.95 & 35.50 & 79.98 & 68.35 & 76.55 & 43.80 & 56.59 & 60.18 \\
\bottomrule
\end{tabular}}
\end{center}
\caption[]{Reasoning evaluation results of our two-phase training approach with \ponebfourptwobone vs. a randomized mixture of both blends across the entire 1T token training run, broken down by individual benchmark. Corresponds to Table \ref{tab:rand-vs-two-phase-result} in \S\ref{subsec:two-phase-results}.
\label{tab:random-BL-eval-results-reasoning}}
\end{table*}

\begin{table*}[t]
\begin{center}
\scalebox{0.72}{\begin{tabular}{@{}lcccccccccc@{}}
\toprule
\textbf{Exp.} & \multicolumn{5}{c}{\textbf{MMLU}} & \multicolumn{5}{c}{\textbf{Code}} \\
\cmidrule(lr){2-6} \cmidrule(lr){7-11}
 & \textbf{STEM} & \textbf{Humanities} & \textbf{Social Sciences} & \textbf{Others} & \textbf{Avg.} & \textbf{HumanEval} & \textbf{HumanEval+} & \textbf{MBPP} & \textbf{MBPP+} & \textbf{Avg.} \\
\toprule
BASE & 50.40 & 49.67 & 66.49 & 63.08 & 56.49 & 28.66 & 25.00 & 44.36 & 44.18 & 35.55 \\
Two-Phase & 49.29 & 50.35 & 67.44 & 62.66 & 56.58 & 31.10 & 24.39 & 49.42 & 43.12 & 37.01 \\
\bottomrule
\end{tabular}}
\end{center}
\caption[]{MMLU and code evaluation results of our two-phase training approach with \ponebfourptwobone vs. a randomized mixture of both blends across the entire 1T token training run, broken down by individual category/benchmark. Corresponds to Table \ref{tab:rand-vs-two-phase-result} in \S\ref{subsec:two-phase-results}.
\label{tab:random-BL-eval-results-MMLU-code}}
\end{table*}

\clearpage
\begin{table*}[t]
\begin{center}
\scalebox{0.64}{\begin{tabular}{@{}llccccccccc@{}}
\toprule
\textbf{Exp.} & \textbf{Tok.} & \textbf{ARC-Easy} & \textbf{ARC-Challenge} & \textbf{RACE} & \textbf{PIQA} & \textbf{WinoGrande} & \textbf{HellaSwag} & \textbf{OpenBookQA} & \textbf{CommonsenseQA} & \textbf{Avg.}\\
\toprule
\ponebfourptwobone & 1T & 80.30 & 54.95 & 35.50 & 79.98 & 68.35 & 76.55 & 43.80 & 56.59 & 60.18 \\
\ponebfourptwobone & 1.7T & 79.84 & 52.90 & 36.27 & 79.65 & 70.01 & 77.79 & 44.80 & 61.75 & 60.88 \\
\ponebfourptwobsix & 1.7T & 80.09 & 55.29 & 37.03 & 80.79 & 70.09 & 78.53 & 46.00 & 60.85 & 61.63 \\
\bottomrule
\end{tabular}}
\end{center}
\caption[]{Reasoning evaluation results of scaling for 1.7T tokens vs. 1T tokens, with and without high-quality data epoch adjustment, broken down by individual benchmark. Corresponds to Table \ref{tab:1.7T-results} in \S\ref{subsec:two-phase-results}.
\label{tab:token-scaling-eval-results-reasoning}}
\end{table*}

\begin{table*}[t]
\begin{center}
\scalebox{0.67}{\begin{tabular}{@{}llcccccccccc@{}}
\toprule
\textbf{Exp.} & \textbf{Tok.} & \multicolumn{5}{c}{\textbf{MMLU}} & \multicolumn{5}{c}{\textbf{Code}} \\
\cmidrule(lr){3-7} \cmidrule(lr){8-12}
 & & \textbf{STEM} & \textbf{Humanities} & \textbf{Social Sciences} & \textbf{Others} & \textbf{Avg.} & \textbf{HumanEval} & \textbf{HumanEval+} & \textbf{MBPP} & \textbf{MBPP+} & \textbf{Avg.} \\
\toprule
\ponebfourptwobone & 1T & 49.29 & 50.35 & 67.44 & 62.66 & 56.58 & 31.10 & 24.39 & 49.42 & 43.12 & 37.01 \\
\ponebfourptwobone & 1.7T & 51.22 & 52.41 & 68.74 & 65.47 & 58.61 & 31.10 & 25.61 & 48.25 & 45.50 & 37.62 \\
\ponebfourptwobsix & 1.7T & 52.39 & 53.41 & 70.20 & 66.91 & 59.85 & 37.20 & 28.66 & 47.08 & 45.50 & 39.61 \\
\bottomrule
\end{tabular}}
\end{center}
\caption[]{MMLU and code evaluation results of scaling for 1.7T tokens vs. 1T tokens, with and without high-quality data epoch adjustment, broken down by individual category/benchmark. Corresponds to Table \ref{tab:1.7T-results} in \S\ref{subsec:two-phase-results}.
\label{tab:token-scaling-eval-results-MMLU-code}}
\end{table*}

\begin{table*}[t]
\begin{center}
\scalebox{0.70}{\begin{tabular}{@{}lccccccccc@{}}
\toprule
\textbf{Model Size} & \textbf{ARC-Easy} & \textbf{ARC-Challenge} & \textbf{RACE} & \textbf{PIQA} & \textbf{WinoGrande} & \textbf{HellaSwag} & \textbf{OpenBookQA} & \textbf{CommonsenseQA} & \textbf{Avg.}\\
\toprule
8B & 80.60 & 53.50 & 37.22 & 80.20 & 70.17 & 76.57 & 45.40 & 61.67 & 61.16 \\
25B & 82.74 & 57.59 & 37.13 & 81.07 & 72.38 & 78.62 & 47.20 & 68.55 & 63.29 \\
\bottomrule
\end{tabular}}
\end{center}
\caption[]{Reasoning evaluation results for 8B vs. 25B parameter models, using the same blend: \ponebfourptwobone, broken down by individual benchmark. Note that we use a maximum sequence length of 8192 (instead of 4096) for both models here. Corresponds to Table \ref{tab:25B-results} in \S\ref{subsec:two-phase-results}.
\label{tab:model-scaling-eval-results-reasoning}}
\end{table*}

\begin{table*}[t]
\begin{center}
\scalebox{0.70}{\begin{tabular}{@{}lcccccccccc@{}}
\toprule
\textbf{Model Size} & \multicolumn{5}{c}{\textbf{MMLU}} & \multicolumn{5}{c}{\textbf{Code}} \\
\cmidrule(lr){2-6} \cmidrule(lr){7-11}
 & \textbf{STEM} & \textbf{Humanities} & \textbf{Social Sciences} & \textbf{Others} & \textbf{Avg.} & \textbf{HumanEval} & \textbf{HumanEval+} & \textbf{MBPP} & \textbf{MBPP+} & \textbf{Avg.} \\
\toprule
8B & 50.05 & 50.82 & 67.70 & 64.21 & 57.31 & 32.93 & 28.66 & 47.47 & 46.83 & 38.97 \\
25B & 58.07 & 59.30 & 77.71 & 72.48 & 65.97 & 37.20 & 33.54 & 58.37 & 53.17 & 45.57 \\
\bottomrule
\end{tabular}}
\end{center}
\caption[]{MMLU and code evaluation results for 8B vs. 25B parameter models, using the same blend: \ponebfourptwobone, broken down by individual category/benchmark. Note that we use a maximum sequence length of 8192 (instead of 4096) for both models here. Corresponds to Table \ref{tab:25B-results} in \S\ref{subsec:two-phase-results}.
\label{tab:model-scaling-eval-results-MMLU-code}}
\end{table*}

\begin{table}[t]
\begin{center}
\scalebox{0.70}{\begin{tabular}{@{}llrrrr@{}}
\toprule
\textbf{Category} & \textbf{Domain} & \textbf{$\mathtt{CMC}$-$\mathtt{B1}$} & \textbf{$\mathtt{CMC}$-$\mathtt{B2}$} & \textbf{$\mathtt{CMC}$-$\mathtt{B3}$} & \\
\midrule
Web Crawl & - & 30 & 15 & 30 \\
\midrule
\multirow{3}{*}{High-Quality} & Math & 23 & 33 & 33 \\
& Wiki & 2 & 2 & 2 \\
& Code & 25 & 30 & 15 \\
\midrule
\multirow{3}{*}{Medium-Quality} & Books & 9 & 9 & 9 \\
& Papers & 11 & 11 & 11 \\
& \ccderiv & 0 & 0 & 0 \\
\midrule
Multilingual & - & 0 & 0 & 0 \\
\bottomrule
\end{tabular}}
\end{center}
\caption[]{Finegrained $\mathtt{CMC}$ \phasetwo Blends (in \%), part 1.}
\label{tab:phase2-finegrained-blends}
\end{table}

\begin{table}[t]
\begin{center}
\scalebox{0.54}{\begin{tabular}{@{}llrrrrr@{}}
\toprule
\textbf{Category} & \textbf{Domain} & $\mathtt{CMC}$-$\mathtt{B3}$-$\mathtt{F10ep}$ & $\mathtt{CMC}$-$\mathtt{B3}$-$\mathtt{F20ep}$ & $\mathtt{CMC}$-$\mathtt{B3}$-$\mathtt{GSM8K}$ & $\mathtt{Combo}$\\
\toprule
Web Crawl & - & 27.1 & 24.2 & 30 & 28.3\\
\midrule
\multirow{3}{*}{High-Quality} & Math & 33& 33 & 31 & 33\\
& Wiki & 2 & 2 & 2 & 2\\
& Code & 15 & 15 & 15 & 15\\
\midrule
\multirow{3}{*}{Medium-Quality} & Books & 9 & 9 & 9 & 9\\
& Papers & 11 & 11 & 11 & 11\\
& \ccderiv & 0 & 0 & 0 & 0\\
\midrule
Multilingual & - & 0 & 0 & 0 & 0\\
\midrule
Task Data & FLAN & 2.9 & 5.8 & 0 & 1\\
& GSM8K & 0 & 0 & 2 & 0.7\\
\bottomrule
\end{tabular}}
\end{center}
\caption[]{Finegrained $\mathtt{CMC}$ \phasetwo Blends (in \%), part 2.}
\label{tab:phase2-finegrained-blends-pt2}
\end{table}

\begin{table}[t]
\begin{center}
\scalebox{0.70}{\begin{tabular}{@{}lrrrrr@{}}
\toprule
\textbf{Blend/Exp.} & \textbf{MMLU} & \textbf{Reason.} & \textbf{GSM8K} & \textbf{Code}	& \textbf{Avg.} \\
\toprule
\phaseone-only & 56.10 & 60.64 & 16.60 & 16.22 & 44.48\\
$\mathtt{CMC}$-$\mathtt{B1}$ & 49.49 & 57.79 & 16.30 & 21.13 & 43.76\\
$\mathtt{CMC}$-$\mathtt{B2}$ & 55.92 & 60.52 & 22.97 & 21.80 & 46.45\\
$\mathtt{CMC}$-$\mathtt{B3}$ & 56.33 & 60.48 & 22.59 & 21.53 & 46.34\\
$\mathtt{CMC}$-$\mathtt{B3}$-$\mathtt{F10ep}$ & 56.45 & 62.80 & 25.70 & 21.46 & 47.89\\
$\mathtt{CMC}$-$\mathtt{B3}$-$\mathtt{F20ep}$ & 56.75 & 62.49 & 26.84 & 22.05 & 47.98\\
$\mathtt{CMC}$-$\mathtt{B3}$-$\mathtt{GSM8K}$ & 56.27 & 60.65 & 35.56 & 21.52 & 47.37\\
\midrule
$\mathtt{CMC}$-$\mathtt{B3}$-$\mathtt{30\%}$ & 56.30 & 59.61 & 32.15 & 23.51 & 47.10\\
$\mathtt{Combo}$ & 56.22 & 62.51 & 45.19 & 25.58 & \textbf{50.27}\\
\bottomrule
\end{tabular}}
\end{center}
\caption[]{Results of finegrained \phasetwo experiments. Code results here average across only HumanEval and MBPP, but not the + versions of both. Hence, they are not directly comparable with the paper results elsewhere.
\label{tab:phase2-finegrained-results}}
\end{table}


\begin{table}[t]
\begin{center}
\scalebox{0.62}{\begin{tabular}{@{}llrrrrr@{}}
\toprule
\textbf{Decay Strategy} & \textbf{Final LR} & \textbf{MMLU} & \textbf{Reason.} & \textbf{GSM8K} & \textbf{Code}	& \textbf{Avg.} \\
\toprule
Linear & $0$ & 56.33 & 61.78 & 22.59 & 21.53 & 46.34\\
Linear & $3e-6$ & 56.16 & 61.63 & 23.35 & 20.68 & 46.08\\
Cosine & $0$ & 56.25 & 61.74 & 21.91 & 21.18 & 46.19\\
Cosine & $3e-6$ & 56.44 & 61.79 & 23.05 & 20.75 & 46.17\\
\bottomrule
\end{tabular}}
\end{center}
\caption[]{Results of different learning rate annealing strategies for \phasetwo.
\label{tab:LR-anneal-results}}
\end{table}

\end{document}